\pdfoutput=1

\documentclass[11pt]{article}

\usepackage{ACL2023}

\usepackage{times}
\usepackage{latexsym}

\usepackage[T1]{fontenc}
\usepackage[utf8]{inputenc}

\usepackage{kotex}
\usepackage{algorithmic}
\usepackage{algorithm2e}
\RestyleAlgo{ruled}
\usepackage{amsmath}
\usepackage{subfigure}
\usepackage{enumitem}
\usepackage[normalem]{ulem}
\useunder{\uline}{\ul}{}
\usepackage{makecell}

\usepackage{xcolor}
\usepackage{graphicx}
\usepackage{multirow}
\usepackage[normalem]{ulem}
\usepackage{color}
\usepackage{vcell}
\usepackage[hang,flushmargin]{footmisc} 
\usepackage{pifont}

\title{RAPID: Training-free Retrieval-based Log Anomaly Detection with PLM considering Token-level information}

\author{Gunho No\thanks{\hspace{2mm}These authors contributed equally to this work.} $^1$ , Yukyung Lee\footnotemark[1] $^1$, Hyeongwon Kang$^1$, Pilsung Kang$^1$ \\
    $^1$Korea University, Seoul, Republic of Korea \\ 
    \texttt{\small $^1$\{gunho\_no, yukyung\_lee, hyeongwon\_kang, pilsung\_kang\}@korea.ac.kr}}

\begin{document}
\maketitle
\begin{abstract}
As the IT industry advances, system log data becomes increasingly crucial. Many computer systems rely on log texts for management due to restricted access to source code. The need for log anomaly detection is growing, especially in real-world applications, but identifying anomalies in rapidly accumulating logs remains a challenging task. Traditional deep learning-based anomaly detection models require dataset-specific training, leading to corresponding delays. Notably, most methods only focus on sequence-level log information, which makes the detection of subtle anomalies harder, and often involve inference processes that are difficult to utilize in real-time. We introduce RAPID, a model that capitalizes on the inherent features of log data to enable anomaly detection without training delays, ensuring real-time capability. RAPID treats logs as natural language, extracting representations using pre-trained language models. Given that logs can be categorized based on system context, we implement a retrieval-based technique to contrast test logs with the most similar normal logs. This strategy not only obviates the need for log-specific training but also adeptly incorporates token-level information, ensuring refined and robust detection, particularly for unseen logs. We also propose the core set technique, which can reduce the computational cost needed for comparison. Experimental results show that even without training on log data, RAPID demonstrates competitive performance compared to prior models and achieves the best performance on certain datasets. Through various research questions, we verified its capability for real-time detection without delay \footnote{Our code is available at {\small \url{https://github.com/DSBA-Lab/RAPID}}}.
\end{abstract}

\section{Introduction}
\label{sec:Introduction}
With the growth of the IT industry, system log data has emerged as an essential resource for diagnosing software incidents, tracing root causes of failures, and early detection of security threats \cite{He2017DrainAO, 7194593, LANDAUER2023100470}.
As software and services become more complex, anomaly detection has become crucial for ensuring the security and stability of the system \cite{9865986}. Hence, log anomaly detection is gaining attention as a technology that can automatically analyze diverse events and patterns in intricate computer systems, identifying unexpected events \cite{10.1145/3510003.3510155}.

\begin{figure*}
  \centering
  \includegraphics[width=0.9\textwidth]{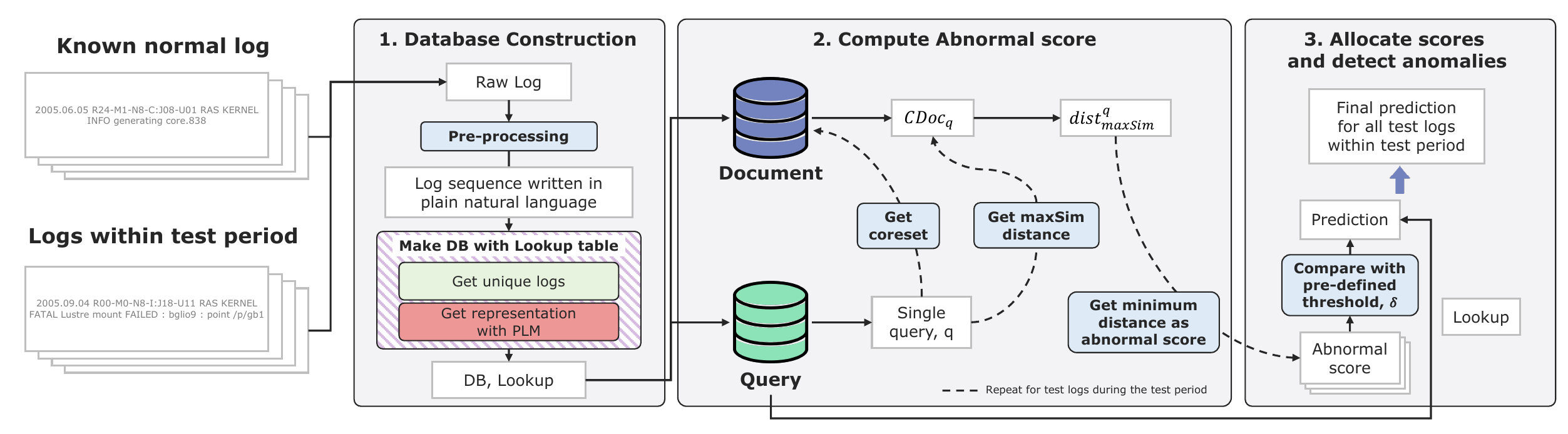} 
  \caption{The overall framework of RAPID. (1) RAPID builds a lookup DB using PLM, updating the Query DB per test period. (2) Each query measures \(\textsf{maxSim}\) distance against the core set of Document DB to determine abnormal. (3) Results are mapped to test timestamps via the Query Lookup.}
  \label{fig:framework}
\end{figure*}

Numerous log anomaly detection methods have been proposed, and most of them employ natural language processing (NLP) for analyzing unstructured system logs. These approaches can mainly be categorized into rule-based, machine learning-based, and deep learning-based algorithms \cite{10.1145/3217871.3217872, 10.1145/3133956.3134015}. Rule-based models identify abnormal logs based on explicit rules and templates \cite{7837916}. Although they are rapid and efficient, rule-based models operate within pre-defined logics, necessitating regular updates \cite{xu2009detecting, LIAO201316}. Conversely, machine learning-based models detect anomalies based on intrinsic patterns trained based on accumulated log data \cite{10.1145/335191.335388,10.1162/089976601750264965,10.1145/1629575.1629587,4781136}. However, the performance of these models depends on feature selection and they exhibit weak adaptability to new log types \cite{10.1145/3338906.3338931}. Recent methodologies leveraging deep learning have demonstrated refined anomaly detection performances using neural networks \cite{10.1145/3133956.3134015, ijcai2019p658, zhang2019robust}. With the proven effectiveness of the Transformer \cite{NIPS2017_3f5ee243}, models like HitAnomaly \cite{huang2020hitanomaly} and LogSy \cite{nedelkoski2020self} were introduced. Moreover, LogBERT \cite{9534113} and LAnoBERT \cite{LEE2023110689}, exploiting the masked language modeling (MLM) approach of BERT \cite{devlin-etal-2019-bert}, have recorded superior detection performances.

Deep learning-based approaches, leveraging various network architectures, have not only effectively handled intricate patterns in log data but also achieved notable performance gains over the years. However, several challenges impede the practicality and scalability of these approaches in real-world scenarios \cite{LANDAUER2023100470, chalapathy2019deep}. First, the development of these models requires training on every log dataset to be utilized, which takes time and therefore delays the start of detection. Furthermore, given that real-world scenarios usually necessitate online updating of accumulated logs, delays also occur in the process of updating the models \cite{10.1145/3501297, cheng2023ai}. Second, most previous studies only focus on sequence-level anomaly detection \cite{9865986}, which may overlook potential benefits from leveraging token-level information within logs \cite{10.1145/3397271.3401075}. Third, many of these models adopt computationally intensive inference processes, such as auto-regressive decoding \cite{10.1145/3133956.3134015, ijcai2019p658, 9534113} or comprehensive token predictions \cite{LEE2023110689}. Such processes can be impractical in situations where millions of log data entries accumulate, potentially making them unsuitable for real-time anomaly detection \cite{cheng2023ai}.

To address these challenges, we introduce RAPID (Training-free \textbf{R}etrieval-based Log \textbf{A}nomaly Detection with \textbf{P}LM cons\textbf{id}ering Token-level information), a novel log anomaly detection model, grounded in three key improvement strategies. First, RAPID employs a retrieval-based reformulation approach, enabled by a pre-trained language model (PLM), to instantiate an anomaly detection methodology capable of identifying anomalous patterns without any training. This methodology facilitates not only the efficient processing of continuously accumulating log data but also significantly enhances the efficacy of delay-free anomaly detection. Second, RAPID amplifies the exploitation of token-level information. Recognizing that even subtle differences or details within log data can serve as crucial indicators for anomaly detection, the model aims to precisely probe into the semantic information of every token, striving to detect abnormal logs with increased performance. Lastly, in consideration of real-time applicability, a concise and fast inference process is designed to enhance the overall procedure. This optimized inference method alleviates complexities, thereby significantly improving the practical applicability of RAPID in real-world environments.

In experiments conducted on BGL \cite{oliner2007supercomputers}, Thunderbird \cite{oliner2007supercomputers}, and HDFS \cite{xu2009detecting} datasets, RAPID achieved notable anomaly detection capabilities. Importantly, without extra training, the model demonstrated competitive results against both supervised and unsupervised models and showed significant performance improvements on complex datasets like BGL and Thunderbird. Furthermore, robust performance was observed in scenarios with limited data and in experiments employing various pre-trained models, demonstrating the potential of RAPID for practical, real-world application.
The main contributions of this study can be summarized as follows:
\begin{itemize}
\item Training-free Detection: The deployment of RAPID with pre-trained language models negates the need for log-specific training.
\item Rich Use of Token-level Information: Enhancement of the capability to capture and utilize token-level details ensures accurate detection of even subtle log discrepancies.
\item Concise Inference: Development of an efficient inference process improves real-world applicability and performance.
\end{itemize}

\section{Related Work}
\label{sec: Related Work}
\subsection{Log-Specific Training Methods}
For anomaly detection, models need to capture the semantic and contextual patterns in log data \cite{LEE2023110689}. Traditional strategies either utilize solely normal log data for model training \cite{10.1145/3133956.3134015, ijcai2019p658} or employ both normal and abnormal logs for binary classification \cite{zhang2019robust, huang2020hitanomaly, 10.1109/ASE51524.2021.9678773, 10.1155/2023/2803139, 10.1016/j.asoc.2022.109860, nedelkoski2020self}. A recent approach leverages the BERT architecture to represent log features through pre-training. LogBERT \cite{9534113} combines MLM with DeepSVDD \cite{pmlr-v80-ruff18a} loss during training and employs a log parser. Additionally, LAnoBERT \cite{LEE2023110689} uses a tokenizer for log sequence pre-processing, eliminating the need for a parser, and emphasizes MLM for training normal log patterns.

Despite their efficacy, the above approaches require training for distinct log datasets, leading to significant training delays. Given that logs contain natural language, PLM can be effectively utilized without additional training, as demonstrated by the robust performance \cite{10.1109/ASE51524.2021.9678773, LEE2023110689}. In this study, we propose an anomaly detection methodology that leverages PLM without requiring log-specific training. Given the rapid accumulation of log data, this methodology offers a practical approach for timely anomaly detection in real-world scenarios. 

\subsection{Feature Embedding Techniques} 
Detecting abnormal patterns in log data requires capturing both semantic and contextual information via feature embedding \cite{10.1145/3510003.3510155}. Such embedding techniques focus on various granularities of logs, utilizing information at either the token or sequence-level. In methods that use token-level information, specific semantic of each log token details are turned into vectors using word embedding \cite{10.1145/3133956.3134015, zhang2019robust, ijcai2019p658, 8109100}. However, focusing on a specific log token may not fully capture the overall log structure or context \cite{9865986}. To address this challenge, studies such as \citet{10.1109/ASE51524.2021.9678773, 9534113} utilized BERT-based contextualized sentence embedding (e.g., [CLS]) to capture both the overall sequence and its global semantic context. Furthermore, \citet{9865986} combined sentence embedding with key event details. 

Our research integrates both token and sequence-level information, enabling the model to capture both detailed semantics and global context within logs. This integrated approach empowers us to accurately and effectively discern the complex structures and varied patterns in logs, potentially enhancing the robustness of anomaly detection performance.

\subsection{Inference Processes} 

The efficacy of the inference process is critical for real-time processing capabilities, especially in environments characterized by continuous data accumulation. Traditional models which can ensure high performance based on complex structures and computations, often encounter challenges when applied in real-world scenarios. Supervised models provide rapid detection in binary classifications; however, their performance may degrade with unseen logs in practical systems \cite{zhang2019robust, huang2020hitanomaly, 10.1109/ASE51524.2021.9678773, 10.1155/2023/2803139, 10.1016/j.asoc.2022.109860}. In contrast, unsupervised model-based approaches commonly employ auto-regressive decoding, which sequentially generates token candidates within logs to identify anomalies \cite{10.1145/3133956.3134015, ijcai2019p658, 9534113}. In addition, LAnoBERT applies a masked auto-encoding technique, utilizing token-level mask probability as an abnormal score \cite{LEE2023110689}. While both approaches exhibit strong detection performance, ensuring their efficiency in computer systems where log data accumulates rapidly remains a significant challenge.

Therefore, this study focused on the importance of efficient inference processes. Our approach leverages general-purpose PLM to derive embeddings from log sequences and conducts retrieval-based anomaly detection. Moreover, we devised an abnormal score that fully encompasses all token-level semantic information in logs, facilitating more accurate anomaly detection by considering even the details of log information that previous methodologies may have overlooked. Significantly, RAPID assures both potent anomaly detection accuracy and adept efficiency in real-time log processing.

\section{Data Analysis}
\label{sec:data_analysis}
\begin{figure}
  \centering
  \includegraphics[width=0.9\columnwidth]{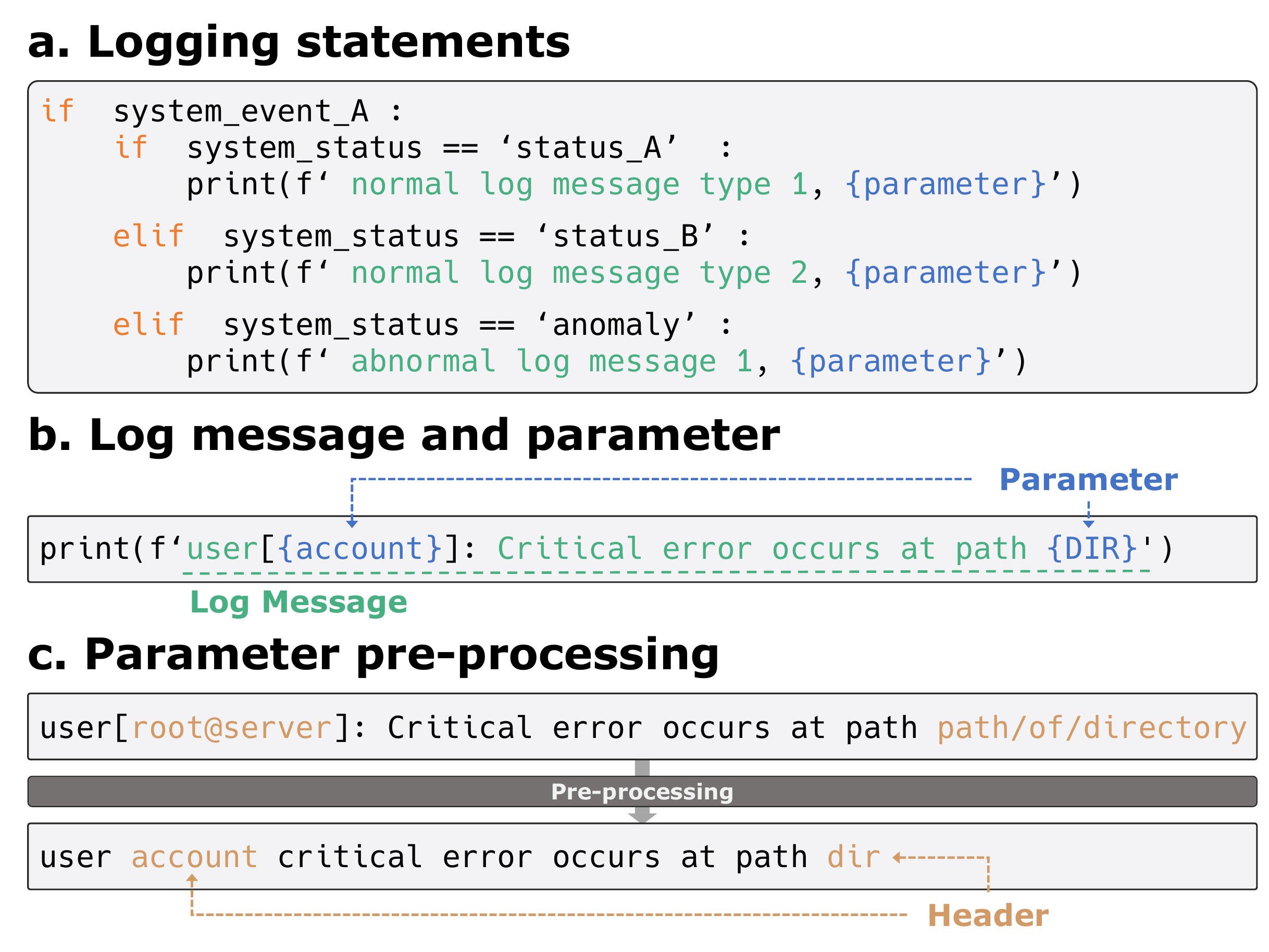}
  \caption{Examples of logging statements in the source code and a format of log data. The figure illustrates log categorization based on system situations, highlighting segments written in natural language and variable parameter segments.}
  \label{fig:log_data}
\end{figure}

\subsection{Log Data} 
\label{sec:log_data}

Logs are textual data that are systematically generated by developers. They use logging statements in the source code to monitor system events and states. The logging statements in Figure~\ref{fig:log_data} (a) show an example in which logs are generated based on specific conditions, including the occurred events and the corresponding system status. Logs are generated within such standardized patterns \cite{Bace2001IntrusionDS, He2017DrainAO}. Figure~\ref{fig:log_data} (b) shows that the log pattern is composed of two parts. One is the log message segment that describes the system situation, and the other is the parameter segment that provides detailed information about specific objects such as `account, directory path, IP'. Notably, logs generated under the same situation share the log message segment, while parameters can vary depending on the instance. Furthermore, log messages are composed of words from natural language in daily use \cite{LEE2023110689}, while parameters consist of words that may have different meanings on different systems \cite{7837916}. For example, there is no guarantee that a particular word in the directory path of a log matches its usage in common language \cite{huang2020hitanomaly}.

Considering these characteristics of logs, we performed minimal pre-processing on the parameter segment to utilize both parts in a unified format for subsequent anomaly detection processes. Due to the structured log generation process, the textual format of parameters representing specific objects can be easily distinguished through regular expressions \cite{10172786}.  Therefore, within each dataset, we replace each parameter token with its corresponding header, such as `account, directory path, IP' within each dataset. This pre-processing is shown in Figure~\ref{fig:log_data} (c).

\begin{figure*}
  \centering
  \includegraphics[width=0.9\textwidth]{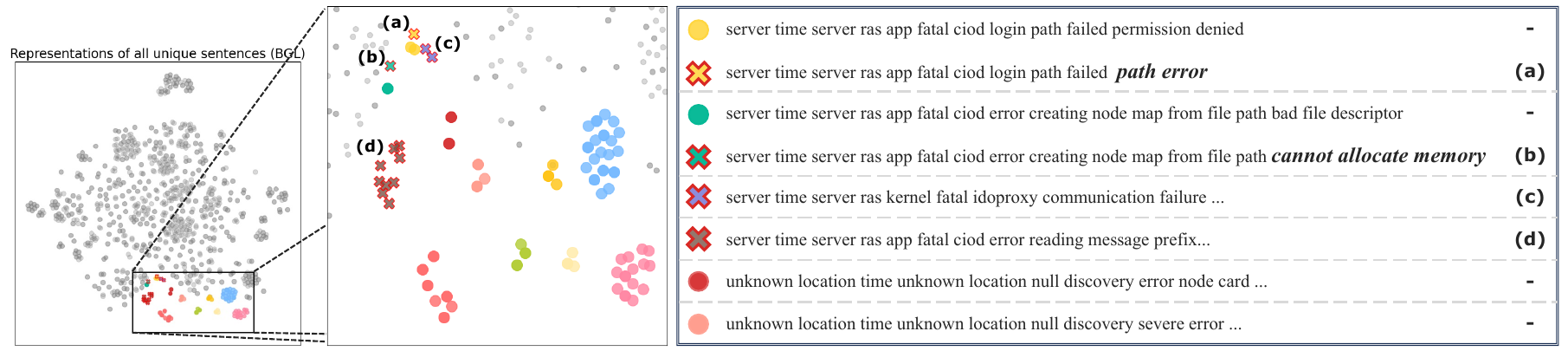}
  \caption{Visualization of representations obtained by passing log sequences from BGL dataset through a PLM (BERT) with t-SNE mapping. It demonstrates that normal logs form clusters, closely interspersed with abnormal logs. When examining the actual log text, the distinctions in described situations are well-reflected in spatial distances.}
  \label{fig:log_dist}
\end{figure*}

\subsection{Log Types} 
\label{sec:log_category}
As observed in Figure~\ref{fig:log_data}, the generation pattern of logs follows the system situation~\footnote{System event and its corresponding status}; therefore logs generated in the same situation exhibit similar structures. Even when the situations are not precisely identical, if the events or statuses are similar, the logs show similarity.  Since logs cover a wide variety of situations, logs might be categorized into various types according to their generation patterns. To validate this, we visualized representations of logs in BGL in Figure~\ref{fig:log_dist} separating normal and abnormal logs. Building upon the insights from the Section~\ref{sec:log_data}, we leveraged representations obtained via a PLM and employed t-SNE \cite{JMLR:v9:vandermaaten08a} to reduce the dimensionality to 2D for visualization. In Figure~\ref{fig:log_dist}, the left side shows all normal log sequences contained within the BGL, while the right side offers an enlarged view, highlighting abnormal log sequences marked with `\ding{54}' symbols and presenting specific examples.

Through the analysis, we have identified three key features in log data. First, various types exist within normal log sequences, and there can be significant differences among these types. This type diversity is caused by the system situation where the logs were generated and is well reflected spatially in Figure~\ref{fig:log_dist}~\footnote{While Figure~\ref{fig:log_dist} represents a 2D reduction from 768-dimensional tensors, distinct log types remain clearly separated.}. Upon analyzing the enlarged figure and the corresponding text of log sequences distinguished by color, log sequences of similar types are positioned closely together in space. This finding means that PLM representation effectively captures the contextual information within the log. Secondly, the distance between normal and abnormal log types can be closer than the distance between different normal log types \cite{7883294}. Even subtle differences, such as a single word, can determine the abnormality, which may lead to abnormal types being remarkably close to normal types. In practice, the abnormal log sequences (a) and (b) in Figure~\ref{fig:log_dist} exhibit text highly similar to the normal type represented by the same color. The distance between these abnormal log sequences and their corresponding normal types was observed to be considerably closer than the distances between different normal types. Third, as shown in (c) and (d) in Figure~\ref{fig:log_dist}, certain abnormal log sequences are positioned among some normal types. This demonstrates the lack of clear distinction between normal and abnormal logs. Considering these observations and noting that each abnormal log sequence is located very close to its corresponding (or similar) normal type, we judge the abnormality of a log for each type. Moreover, we design anomaly detection method based on the type each test log sequence belongs to, rather than using a single normal criterion.

\subsection{Token-level Information}\label{sec:log_token} 

Another characteristic of the log is that it has a very limited vocabulary size compared to the vast data size \cite{10.1145/3133956.3134015}. In fact, the vocabulary size of BGL, Thunderbird, and HDFS, the three representative data used as benchmarks, are 888, 3137, and 229, respectively, which are quite small compared to the diversity of logs. This suggests that anomaly detection using token-level information can be advantageous. Since anomaly detection solely relies on known normal log sequences \footnote{Known normal log sequence is the same as the train normal log in previous studies}, the key to performance is how effectively new incoming logs can be interpreted based on these known normal sequences. Intuitively, if we judge anomalies by comparing test logs to known normal logs, the quality of this interpretation can be inferred by how well the known normal log sequence contains information about unseen test logs. To evaluate the coverage of this information at the sequence-level and token-level, we examine the ratio of seen test log sequences and seen test tokens via known normal logs, respectively. In the former case, we define a test log sequence as covered if it is equally present in the known normal log sequence, and in the latter case, coverage is calculated by how much each token appearing in the test log is included in the vocabulary appearing in the entire known normal log sequence.

\begin{figure}
  \centering
  \includegraphics[width=\columnwidth]{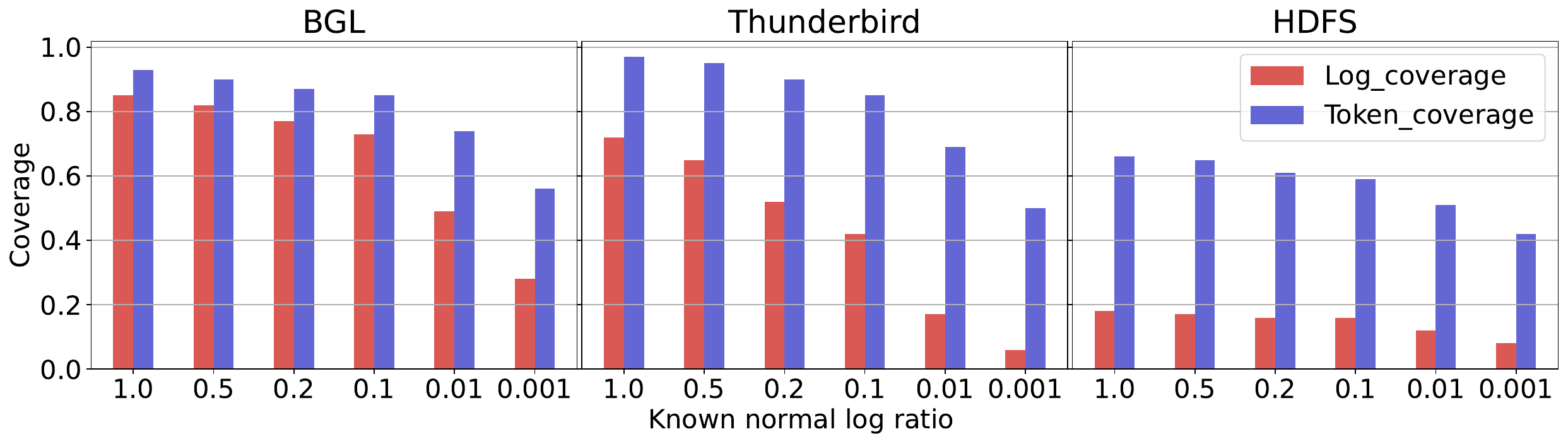}
  \caption{Illustration comparing log sequence coverage of a known normal log sequence for a test log with token coverage. It shows that the interpretation of the test log is more advantageous when the information in the token is utilized.}
  \label{fig:token_coverage}
\end{figure}

As shown in Figure~\ref{fig:token_coverage}, test log token coverage of known normal log sequences is higher than test log sequence coverage. These results suggest that more detailed insights into test logs can be gained by directly considering token-level information within the logs, rather than relying solely on sequence-level information. Furthermore, in simulations where logs are accumulated by utilizing only a certain percentage of the known normal log sequence, it is observed that even when the amount of available known normal log sequences decreases, token coverage has a relatively larger value compared to log sequence coverage. This also suggests that utilizing all tokens can provide a more precise interpretation of test log information.

\section{proposed method}
\subsection{Task Formulation}
\subsubsection{Traditional Log Anomaly Detection Task Formulation}

Traditional log anomaly detection aims to evaluate the normality of a given log sequence. In particular, a log sequence that differs from the normal pattern observed in the training data is considered abnormal, and usually only normal data is utilized for training. A log sequence $\mathcal{L}$ is represented by the following equation (\ref{eq:L}).
\begin{align}
\mathcal{L} = \{t_1, t_2, \ldots, t_{N}\} \label{eq:L}
\end{align}
$t_1$ represents the $i$-th token and $N$ denotes the sequence length of $\mathcal{L}$. Equation (\ref{eq:s_ad}) indicates that a machine learning model $f$, trained solely on normal log data, is employed to map the input $\mathcal{L}$ to the abnormal score function $s_{ad}$, where $\theta$ represents the parameters the of model.
\begin{align}
s_{ad}(\mathcal{L}; \theta) = f(\mathcal{L}) \label{eq:s_ad}
\end{align}
The sequence $\mathcal{L}$ is detected as abnormal if $s_{ad}(\mathcal{L}; \theta) \geq \delta$ according to the Equation~(\ref{eq:abnormal}), where $\delta$ is the threshold. 
\begin{align}
\label{eq:abnormal}
\text{Anomaly} =
\begin{cases}
    1 & \text{if}~s_{ad}(\mathcal{L}; \theta) \geq \delta \\
    0 & \text{otherwise}
\end{cases}
\end{align}

\subsubsection{Retrieval-based Reformulation}
Based on the analysis results in Section \ref{sec:data_analysis}, we propose an information retrieval (IR) based reformulation to design training-free log anomaly detection. Specifically, we introduce the anomaly detection process centered on the relationship between known normal log sequences and new test data. For a test log that contains both normal and abnormal logs, the proposed approach involves finding a normal log corresponding to each test log, and judging it as abnormal when the distance between the pair is larger than the threshold. More precisely, each test log sequence takes on the role of a query, while each known normal log sequence act as a document in IR. In other words, the abnormal score for this query $q$ is calculated through the similarity with the target document $d$ as shown in the Equation~(\ref{eq:rb}).
\begin{align}
\label{eq:rb}
s_{ad}(q) = 1 - g(q, d)
\end{align}

Here, the function $g$ represents the similarity between the two log sequences and uses the cosine similarity score of PLM representations. As shown in Equation~(\ref{eq:abnormal}), if $s_{ad}(q) < \delta$, the test log sequence $q$ is judged to be normal because it is similar to a known normal log sequence, and vice versa. Through the described reformulation, we can perform anomaly detection by simply comparing the distance between test logs and normal log sequences without any additional training.

\subsection{RAPID}
\label{sec:RAPID}
In this section, we provide a detailed explanation of the RAPID process based on the proposed retrieval-based log anomaly detection (R-LogAD) formulation, specifically explaining how each step overcomes the limitations of previous research. The key points of RAPID are summarized as follows. First, RAPID analyzes log sequences using a general-purpose PLM and performs R-LogAD, which eliminates the need for training. Second, it leverages feature embedding that actively reflects all token semantic information in the log sequence to include as much information as possible. Third, anomaly detection operations are performed only on the type to which the test log belongs, resulting in an efficient and concise inference process.

\subsubsection{Feature Embedding}
The first step is to get the feature embedding from the log, and this step builds the query database ($Q$) and document database ($D$).

\textbf{Parameter pre-process with regular expression}
Detection process starts with log pre-processing. As pointed out in \cite{zhu2019tools}, log parsers used in many previous studies \cite{huang2020hitanomaly, 10.1145/3338906.3338931, 10.1109/NOMS54207.2022.9789917, 9865986, 10.1155/2023/2803139, 7883294,10.1145/3133956.3134015,ijcai2019p658,9534113} do not always work correctly on all log datasets. Furthermore, the use of such parsers may result in the loss of semantic information. In contrast, the proposed regular expression-based pre-processing described in Section~\ref{sec:log_data}, despite its simplicity, provides performance advantages \cite{10.1016/j.asoc.2022.109860, 10.1109/ASE51524.2021.9678773}. The log sequences obtained after pre-processing contain both log messages and parameters expressed in common language text with minimal information loss. This composition enables effective utilization of general-purpose PLMs in anomaly detection.

\textbf{Make unique log sequence database}
Performing parameter pre-processing on a log dataset unifies the distinctive parameters for each individual log instance, so logs generated in the same situation will be redundant \cite{LEE2023110689}. Therefore, the proposed R-LogAD only requires a single document log sequence for each situation. Using this approach, RAPID constructs database $D$ using only unique normal log sequences, significantly reducing the retrieval search space. The same approach can be applied to the test logs used as queries. As logs accumulate at a fast rate, typically at 5 times or more per second \cite{oliner2007supercomputers}, even real-time detection is conducted for a certain testing period at a time. Therefore, RAPID constructs database $Q$ using unique test log sequences and calculates abnormal scores within the test period. It then assigns scores to their corresponding actual test timestamps through a database lookup table. This process significantly reduces inference time by decreasing the frequency of the detection process, and the benefit increases as the test period increases.

In anomaly detection, especially when abnormal log information is unavailable, understanding the intent of the developer embedded in the logs becomes crucial. This suggests that detection should factor in the semantic information present within the log text \cite{10.1109/ASE51524.2021.9678773, 9865986, 10.1016/j.asoc.2022.109860, LEE2023110689}. Therefore, after constructing databases $D$ and $Q$, the embedding of included log sequences is added to the database using PLM. In the case of $D$, we continue to use the stored results after one encoding process, and in the case of $Q$, we encode them every test period. Since both databases are already built with the minimum required log sequence, this process is very efficient in terms of computational cost. Also, by directly utilizing the required document information for queries from database $D$, it significantly contributes to efficient inference. A more detailed description of the feature embedding extraction process can be found in Algorithms~\ref{alg:DB}.

\begin{algorithm}
\DontPrintSemicolon 
\KwData{$L$, $L^{\prime}$: Set of logs and processed log sequences,  \\ 
$DB$: Database of unique log sequences,\\ $Lookup$: Database of index by timestamp, \\
$idx$: Index of unique log sequence in $DB$, \\
\textsf{regex}: Regular expression pre-processing function, \\
$Seq$: Key of DB indicating $L^{\prime}$, \\$E$: Key of DB indicating embedding of $L^{\prime}$.}
\KwResult{$DB$, $Lookup$}

\BlankLine
\tcp{Initialization}
$idx \leftarrow 0$\;
\BlankLine
\tcp{Database Construction}
\For{$i=1$ \KwTo \textsf{len}($L$)}{
    $L^{\prime}_{i} \leftarrow \textsf{regex}(L_{i})$\;
    \If{$L^{\prime}_{i}$ not in $DB[\text{Seq}]$}{
        $idx \leftarrow idx + 1$\;
        $Lookup[i] \leftarrow idx$\;
        $DB[idx][\text{Seq}] \leftarrow L^{\prime}_{i}$\;
        $DB[idx][\text{E}] \leftarrow \textsf{PLM}(L^{\prime}_{i})$\; 
    }
}
\Return{$DB$, $Lookup$}\;
\caption{Database Construction}\label{alg:DB}
\end{algorithm}

\subsubsection{Selection of the Comparison Log Sequence for Each Query from the Documents}
\label{sec:comparison_target}
The key to R-LogAD is to select an appropriate document log sequence from $D$ to judge each query log sequence. Especially, since RAPID is a training-free method, it needs a comparison target that can determine the normality only using the representation of the log sequence itself. Based on the analysis of prior work by \citet{7883294} and Section~\ref{sec:log_category}, the selection of a comparison target should be determined according to the type of query. This perspective is also found in other anomaly detection studies besides logs, consistent with the claim that different clusters exist within normal data \cite{you2022a}. However, which type the query log sequence belongs to, or even whether that type exists in $D$ is not known. Since the log sequence representation obtained through PLM contains context information about the system situation, logs of similar situations will be located close in the embedding space \cite{devlin-etal-2019-bert}. Leveraging this property, it is assumed that the nearest document type to a query, based on the log sequence representation, is the same or a similar log type as that query. However, when performing anomaly detection under this assumption, it is challenging to determine the number of log sequences to include in the nearest document type for abnormal score calculation, as it is unknown. Including more log sequences than the actual number results in an incorrect score that reflects unrelated types. On the other hand, using fewer log sequences would not capture other types. Therefore, RAPID intuitively performs anomaly detection by comparing to the single nearest document log sequence. Verification of this approach is performed in Section~\ref{sec:nearest_category}.

\subsubsection{Token-level Information Based Max Similarity Distance}

RAPID performs anomaly detection by comparing each query log sequence with its nearest document log sequence. This approach requires a clear definition of the criterion for measuring distance and the method for comparing that distance. In this context, the distance between log sequence representations is selected to calculate abnormal score. In particular, while previous studies mainly utilized only sequence-level information, we applied \textsf{maxSim} distance inspired by ColBERT to actively reflect token-level information. The \textsf{maxSim} score computes the query-document similarity by comprehensively reflecting the cosine similarity between all tokens and following the equation (\ref{eq:maxsim}) for embedding $E$. 
\begin{equation}
\label{eq:maxsim}
\textsf{maxSim}(q, d) = \sum_{i \in\left[\left|E_q\right|\right]} \max _{j \in\left[\left|E_d\right|\right]} E_{q_i} \cdot E_{d_j} .
\end{equation}
where $\cdot$ represents cosine similarity function, $q_i$ and $d_j$ are tokens in $q$ and $d$ respectively. Because the log sequence representation obtained through PLM has both the CLS token embedding and the embedding of each word token, the \textsf{maxSim} operation can reflect both the sequence-level and the token information \cite{10.1145/3397271.3401075}. After that, following the formula \text{{distance}} = 1 - \text{{similarity}}, we reverse the sign to define the query-document \textsf{maxSim} distance as in Equation (\ref{eq:distance}).
\begin{equation}
\label{eq:distance}
\textsf{distance}(q, d) = 1 - \textsf{maxSim}(q, d) .
\end{equation}
Finally, the abnormal score for each query log sequence is determined by the smallest \textsf{maxSim} distance between that query log sequence and all log sequences in $D$.

\subsection{Algorithm} 
This section details how RAPID performs anomaly detection using the database $D$, $Q$, comparison target document for each query, and \textsf{maxSim} distance defined in Section~\ref{sec:RAPID}.

\subsubsection{Get Core Set}
By our definition, the abnormal score for a query log sequence is determined by its \textsf{maxSim} distance from the nearest document log sequence. The challenge with this process is that determining the `nearest' document log sequence requires computing the \textsf{maxSim} distance to all log sequences in $D$. This is quite inefficient given that only the distance to the single document log sequence is ultimately used for anomaly detection. In particular, the \textsf{maxSim} calculation involves computing the cosine similarity between all tokens, which is very computationally expensive. To tackle this challenge, we introduce a process to extract the core set so that only a minimal number of \textsf{maxSim} calculations can be performed.

To extract the core set, we focus on the CLS token, which summarizes sequence-level information. This special token contains sufficient information to select the nearest document candidates as it has been directly utilized for anomaly detection in many previous studies \cite{huang2020hitanomaly, 10.1145/3338906.3338931, 10.1109/ASE51524.2021.9678773}.
As a result, only the Euclidean distance between the CLS tokens of the query log sequence and the document log sequences is required. Nevertheless, it cannot be guaranteed that the document obtained through the CLS Euclidean distance will always match that obtained using the `maxSim' distance, which considers all tokens. Therefore, to account for this uncertainty, we implemented the K-nearest neighbor (KNN) algorithm for this computation, selecting the pre-specified K neighboring document log sequences as candidates. The process of determining the core set $CDoc_{q}$ for a specific query log sequence $q$ is as follows.

\begin{align}
CDoc_{q} &:= \textsf{Core}(q, D, K), \label{eq:Cdoc} \\
\textsf{Core}(q, D, K) &= KNN(q, D), \label{eq:Core} \\
\begin{split}
KNN(q, D) &= \left\{ d_i \in D \, \middle| \, i \in \textsf{top-K}\right. \\
&\quad \left. \left(-\| E_{q}^{CLS} - E_{d_i}^{CLS} \|_2\right) \right\}.
\end{split}
\label{eq:KNN}
\end{align}

As shown in the Equation (\ref{eq:Cdoc}), the core set is selected via the function \textsf{Core} defined in (\ref{eq:Core}). The calculation through the KNN algorithm is efficiently executed, yielding top-k nearest neighbors as candidates, as defined by Equation (\ref{eq:KNN})
\begin{figure}
  \centering
  \includegraphics[width=0.95\columnwidth]{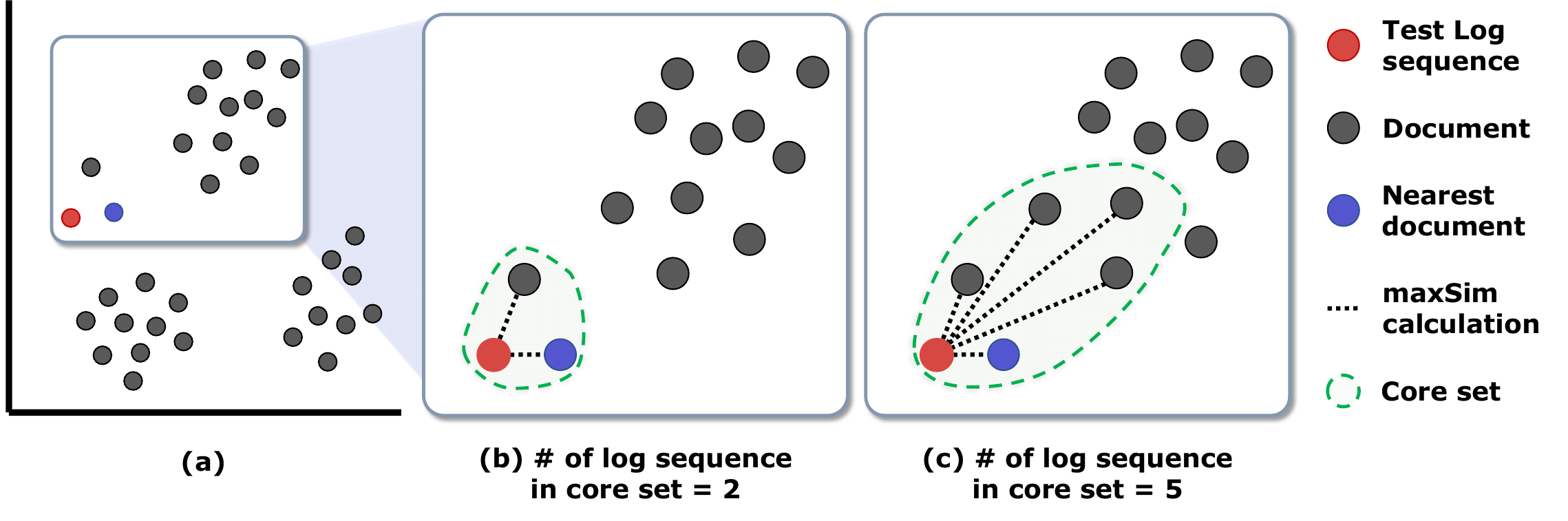}
  \caption{(a) Spatial representation of documents (known normal log sequence) and a query (test log sequence) based on the \textsf{maxSim} distance. (b) and (c) show partial views with core sets of 2 and 5, respectively.}
  \label{fig:coreset}
\end{figure}

Utilizing a core set means that the \textsf{maxSim} distance is only calculated for the documents in the selected core set of query, regardless of how many documents there are in total. In (b) and (c) of Figure~\ref{fig:coreset}, the distance calculation is only performed for documents within the dashed area. Finally, the shortest of these distances is taken as the abnormal score.

\subsubsection{Abnormal Score}
The process of calculating an abnormal score in RAPID can be summarized as follows, and a detailed description can be found in Algorithms~\ref{alg:AD}.
\begin{enumerate}
    \item Database construction:
    Using regular expressions, replace the parameter part of the log with the header of the corresponding object, and generate $D$ and $Q$ with only unique log sequences. Additionally, only those log sequences contained in this database will be represented by PLM.
    \item Compute abnormal score:
    For an efficient inference step that can be used in a real-world anomaly detection scenario, we apply a KNN algorithm utilizing only the representation of CLS tokens to narrow the search space for each query. We perform the \textsf{maxSim} distance operation on this core set only and determine the minimum value as the abnormal score for that query.
    \item Allocate scores and detect anomalies:
    To convert the anomaly score calculated based on $Q$ into the result for the input test log, detection is performed on this score using the defined threshold. Then, predictions are assigned to each timestamp in the test period through a lookup table created with the $Q$.
\end{enumerate}

\begin{algorithm}
\DontPrintSemicolon 
\KwData{$L_{kn}$: Known normal logs,\\$L_{ts}$: Test logs during detection, \\
$D$: Database of known normal logs, 
\\$Q$: Database of test logs, \\
$CDoc_i$: Sub-set of $Document$ for $i^{th}$ query, \\
$dist_{maxSim}$: Calculated \textsf{maxSim} distance, \\
$\delta$: Anomaly detection threshold.}
\KwResult{$final\_pred$}
\BlankLine
\tcp{Make database}
$D, Lookup_D \leftarrow \textsf{Make\_DB}(L_{kn})$\;
$Q, Lookup_Q \leftarrow \textsf{Make\_DB}(L_{ts})$\;
\BlankLine
\tcp{Compute Abnormal score}
\For{$i=1$ \KwTo \textsf{len}($Q$)}{
    $CDoc_{i} \leftarrow \textsf{Core}(Q[i], D, K)$\;
    \For{$j=1$ \KwTo K}{
        $dist_{maxSim}^i \leftarrow 1-\textsf{maxSim}(Q[i], CDoc_{i}[j])$\;
    }
    $Abnormal\_score[i] \leftarrow \textsf{min}(dist_{maxSim}^i)$\;
}
\BlankLine
\tcp{Allocate scores and detect anomalies}
\For{$k=1$ \KwTo \textsf{len}($Lookup_Q$)}{
    \eIf{$Abnormal\_score[Lookup_Q[k]] \geq \delta$}{
        $final\_pred[k] \leftarrow \text{Abnormal}$\;
    }{
        $final\_pred[k] \leftarrow \text{Normal}$\;
    }
}
\Return{$final\_pred$}\;
\caption{RAPID Processing}\label{alg:AD}
\end{algorithm}

\begin{table}
\centering
\resizebox{\columnwidth}{!}{%
\setlength{\tabcolsep}{6pt}
\renewcommand{\arraystretch}{1.4}
\begin{tabular}{lccccc}
\Xhline{3\arrayrulewidth}
& \textbf{Train} & \textbf{Parser} & \textbf{BGL}    & \textbf{Thunderbird} & \textbf{HDFS} \\
\hline
\textit{Supervised} & ~ & ~ & ~ & ~ & ~ \\
\hline
\textbf{LogRobust} & \ding{51} & \ding{51} & 0.8300 & - & 0.9700 \\
\textbf{HitAnomaly} & \ding{51} & \ding{51} & 0.9200 & - & 0.9800 \\
\textbf{LogSy} & \ding{51} & \ding{55} & 0.6500 & {\ul \textbf{0.9900}} & - \\
\textbf{NeuralLog} & \ding{51} & \ding{55} & 0.9800 & 0.9600 & 0.9800 \\
\textbf{Adanomaly} & \ding{51} & \ding{51} & 0.9200 & - & 0.9800 \\
\textbf{DeepSyslog} & \ding{51} & \ding{51} & 0.9800 & - & 0.9800 \\
\textbf{LogPal} & \ding{51} & \ding{51} & {\ul \textbf{0.9900}} & \textbf{0.9800 }& {\ul \textbf{0.9900}} \\
\textbf{LayerLog} & \ding{51} & \ding{55} & \textbf{0.9850} & - & \textbf{0.9880} \\
\hline
\textit{Unsupervised} & ~ & ~ & ~ & ~ & ~ \\
\hline
\textbf{PCA$^\dagger$} & \ding{51} & \ding{51} & 0.1661 & 0.5439 & 0.1112 \\
\textbf{iForest$^\dagger$} & \ding{51} & \ding{51} & 0.3065 & 0.0329 & 0.6049 \\
\textbf{OCSVM$^\dagger$} & \ding{51} & \ding{51} & 0.0196 & 0.2548 & 0.0495 \\
\textbf{LogCluster$^\dagger$} & \ding{51} & \ding{51} & 0.7663 & 0.5961 & 0.5399 \\
\textbf{DeepLog$^\dagger$} & \ding{51} & \ding{51} & 0.8612 & 0.9308 & 0.7734 \\
\textbf{LogAnomaly$^\dagger$} & \ding{51} & \ding{51} & 0.7409 & 0.9273 & 0.5619 \\
\textbf{LogBERT$^\dagger$} & \ding{51} & \ding{51} & \textbf{0.9083} & 0.9664 & 0.8232 \\
\textbf{LAnoBERT} & \ding{51} & \ding{55} & 0.8749 & {\ul \textbf{0.9990}} & {\ul \textbf{0.9645}} \\
\textbf{RAPID (ours)} & \ding{55} & \ding{55} & {\ul \textbf{0.9999}} & \textbf{0.9975} & \textbf{0.9240} \\
\Xhline{3\arrayrulewidth}
\end{tabular}}
\caption{F1-score on BGL, Thunderbird, and HDFS. ${}^\dagger$ indicates the performance of benchmark models reported by LogBERT. Best performance is bolded and underlined, while second-best is only bolded. (RQ1)} \label{tab:main}
\end{table}

\section{Research Questions and Results}
The detailed experiment setup including dataset and baselines, and additional research questions can be found in Appendix~\ref{sec:exp_setup},~\ref{sec:additional_RQ}.
\subsection{Does the RAPID achieve outperform anomaly detection performance? (RQ1)}\label{sec:main_result}
We validate the performance of RAPID by comparing it to previous studies with log-specific training. Table~\ref{tab:main} compares the performance of RAPID and the baseline models, categorized by whether each method is supervised or trained, whether token information is utilized in the anomaly detection process, and whether a log parser is used. RAPID is the only model that does not require any training.

In the comparison with the supervised setting models, RAPID shows competitive performance. Specifically, NeuralLog is a methodology that utilizes a general-purpose PLM similar to ours but focuses on binary classification. Even in this comparison, our model performs better, indicating that our methodology effectively utilizes the representation of the PLM to perform anomaly detection without abnormal data.

When comparing the performance in the unsupervised setting, the proposed RAPID achieved strong performance on all datasets. It shows the best detection performance on the BGL dataset, even with the supervised learning setting, and the second-best performance on the Thunderbird dataset after LAnoBERT. It even performs quite competitively on the HDFS dataset, even though the method is designed for a single log input. This robust performance supports the effectiveness of the abnormal score designed by RAPID. Compared to methodologies that consider the type of logs, such as LogCluster, it outperforms all datasets. In addition, when we examine LogBERT and LAnoBERT, which use a well-known PLM, BERT, alongside RAPID, it becomes clear that the PLM-based model shows higher detection performance overall. This shows the excellent capability of the PLM to analyze log data and suggests that even higher performance can be expected because the PLM can efficiently leverage token information from the representation of logs. In particular, unlike the two previous studies, our model does not train PLM on logs at all, yet it performs as well as or better than the two previous models. This confirms that the proposed RAPID utilizes the PLM representation very effectively.

\subsection{Is it valid to select only the one document log sequence nearest to the query as a comparison target? (RQ2)}\label{sec:nearest_category} 
In Section~\ref{sec:comparison_target}, we defined anomaly detection for a test log by comparing its query to the nearest normal type within $D$. The objective was to evaluate the type to which the query most likely belongs. Since the number of log sequences belonging to that normal type is unknown, we substituted it with a comparison to the single nearest document log sequence. To evaluate the proposed approach, we compare the performance of considering all log sequences in the core set and only the single nearest log sequence to the query, for different core set sizes.
    
In Table~\ref{tab:nearest_category}, based on the query log sequence, we compare the performance of using only the distance to the nearest single log sequence in the core set as the abnormal score (nearest only) and using the average value of the distance to all log sequences in the core set as the ad score (core set mean). For the three datasets, the performances are recorded by varying the number of log sequences in the core set, starting with the case where the number of log sequences is 2 because the average calculation of a single distance is meaningless.

First, we validate the anomaly detection approach using only the nearest normal type to the query log. The `ALL' row refers to the case where distance is calculated for the entire $D$, and the `nearest only' and `core set mean' columns show the detection performance of anomaly detection using only the nearest normal type and anomaly detection reflecting all normal data, respectively. In the BGL and Thunderbird datasets, the F1-scores for `nearest only' were 0.9999 and 0.9979, respectively, whereas for `core set mean' they were 0.4659 and 0.7586. This confirms that the performance when using only the nearest type is significantly superior. This shows that as proposed in Section~\ref{sec:comparison_target}, the approach of using only the nearest normal type to the query log during detection is valid. In the case of HDFS, which is a block-based dataset consisting of multiple log sequences as described earlier, performance shows a different pattern because each query and document does not refer to a single log sequence. Nevertheless, it outperforms the competing baseline models by only utilizing the nearest type of document.

Next, we validate the approach of reflecting only the nearest single document log sequence instead of the normal type. Comparing the performance of the `nearest only' and `core set mean' columns for BGL, we can see that as the number of log sequences in the core set changes from 2, 5, and 10, the F1-score is robust at 0.9999 when using only the single nearest log sequence, while the performance changes significantly and drops up to 0.5127 when including all log sequences in the core set. The same trend is observed for Thunderbird. This results due to the number of log sequences forming the core set actually exceeding the number of log sequences contained in the nearest type to the query, which may cause performance degradation as anomalies are determined from other types unrelated to the query. As mentioned, the performance of `core set mean' in the BGL dataset notably decreases as the number of log sequences in the core set increases from 2 to 5. This case is illustrated in Figure~\ref{fig:coreset} (b) and (c). Since the `Nearest only' performance is robust across various core set sizes, in conclusion, anomaly detection by comparing a test log sequence to the single nearest log sequence is valid.

\begin{table}
\centering
\resizebox{0.95\columnwidth}{!}{%
\setlength{\tabcolsep}{5pt}
\renewcommand{\arraystretch}{1.4}
\begin{tabular}{c|c|c|c} 
\Xhline{3\arrayrulewidth}
\multirow{2}{*}{\textbf{Dataset}}                             & \multirow{1}{*}{\textbf{ \# of log sequence}} & \multicolumn{2}{c}{\textbf{F1}}           \\ 
\cline{3-4}
            & \multirow{1}{*}{\textbf{in core set}}                     & \textbf{core set mean} & \textbf{nearest only}  \\ 
\hline
\multirow{4}{*}{\textbf{BGL}}         & 2                                      & 0.9689                & 0.9999                  \\
                             & 5                                               & 0.7299                & 0.9999                  \\
                             & 10                                              & 0.5127                & 0.9999                 \\
                             \cline{2-4}
                             & ALL                                             & 0.4659                & 0.9999                  \\ 
\hline
\multirow{4}{*}{\textbf{Thunderbird}} & 2                                      & 0.5073                & 0.9978                   \\
                             & 5                                               & 0.5641                & 0.9978                  \\
                             & 10                                              & 0.7989                & 0.9979                  \\
                             \cline{2-4}
                             & ALL                                             & 0.7586                & 0.9979                  \\ 
\hline
\multirow{4}{*}{\textbf{HDFS}}        & 2                                      & 0.9333                & 0.9240                 \\
                             & 5                                               & 0.9485                & 0.9240              \\
                             & 10                                              & 0.9700                & 0.9240               \\
                             \cline{2-4}
                             & ALL                                             & 0.9495                & 0.9240                \\
\Xhline{3\arrayrulewidth}
\end{tabular}}
\caption{Performance based on the number of log sequences in the core set and those included in the abnormal score calculation. `Core set mean' considers all sequences equally, while `nearest only' looks at the nearest log sequence. (RQ2)}
\label{tab:nearest_category}
\end{table}

\begin{figure*}
  \centering
  \includegraphics[width=\textwidth]{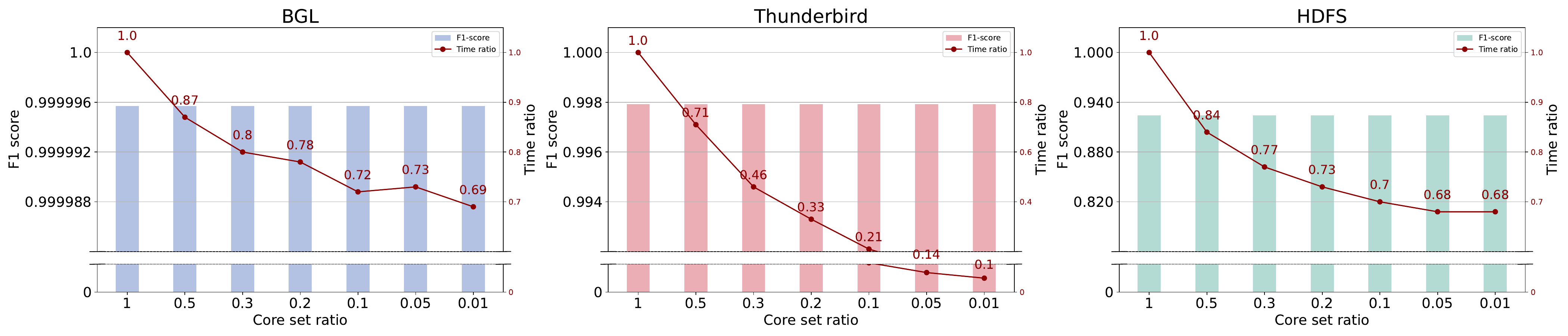}
  \caption{F1-score and inference time by a ratio of the utilized core set in entire $D$. The reduction in the size of the core set used for \textsf{maxSim} calculation maintains the performance while significantly decreasing the inference time. (RQ3)}
  \label{fig:coreset_ratio}
\end{figure*}

\subsection{Does RAPID have an efficient inference process for real-world applications? (RQ3)}\label{sec:real-world}
To assess inference efficiency, we compare the performance and inference time using different size core sets. Figure~\ref{fig:coreset_ratio} shows the performance and time taken by the model according to the size of the core set, where the x-axis indicates the ratio of log sequences used as a core set to the total $D$. The line graph shows the proportion of inference time spent at each core set size, with 1 indicating the case where computing the \textsf{maxSim} distance for all log sequences in $D$ without core sets. For all three benchmark datasets, we can see that the detection performance of the model, as visualized by the bar charts, remains unchanged even as the size of the core set decreases. However, the inference latency decreases significantly. This demonstrates that our proposed core set technique enables a substantial improvement in inference speed while maintaining performance. Particularly in the case of the larger dataset, Thunderbird, the inference time is reduced by up to ten times. These findings indicate that our methodology can maintain a remarkably fast inference speed even as data accumulates and the number of known normal log sequences expands over time. For instance, in a scenario where $D$ is already constructed and using only 0.01 of the total known normal log sequences as a core set for detection, we can process about 12,000 and 3,000 test logs per second for BGL and Thunderbird, respectively. Considering that each data set generates about 11 and 8 log outputs per second, respectively, this demonstrates that our model has an efficient inference process.

\subsection{Can RAPID detect anomalies without delay under real-time data accumulation scenarios? (RQ4)}
\label{sec:real_time}
\begin{figure}
  \centering
  \includegraphics[width=\columnwidth]{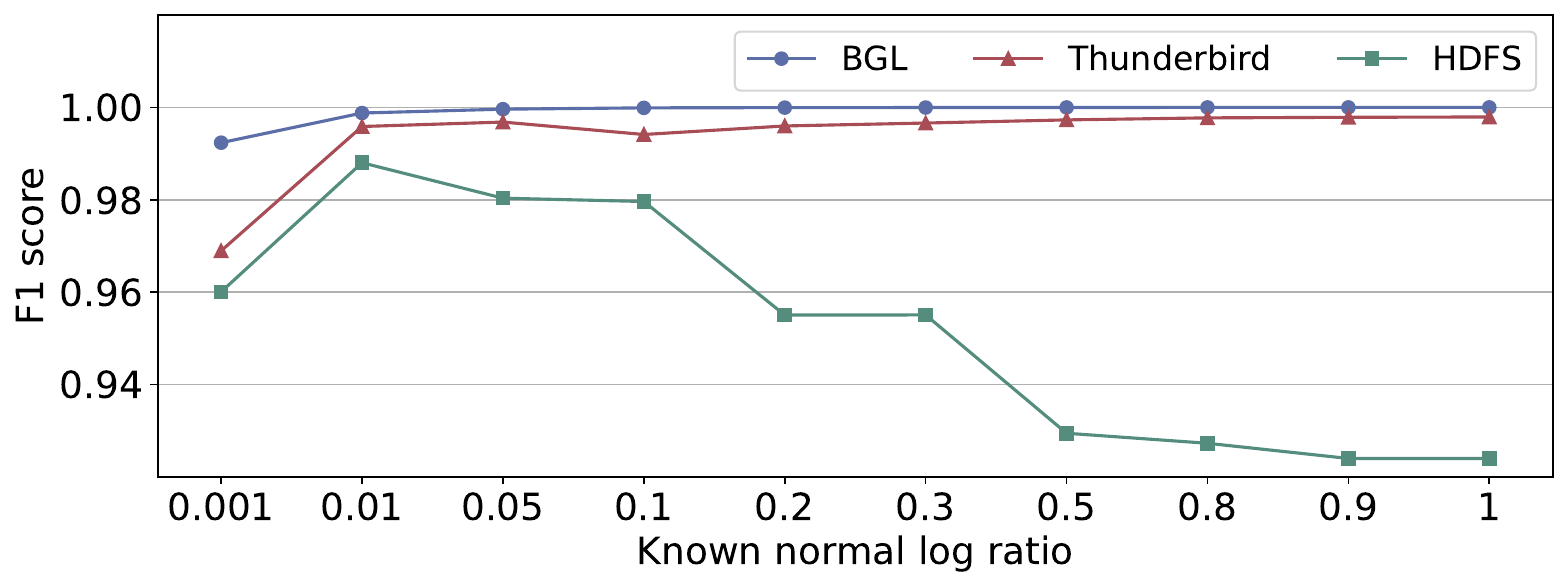}
  \caption{F1-score by using known normal log ratio. Reducing the portion of the known normal log, simulating a real-time data accumulation scenario, demonstrates robust performance even with very few known normal logs. (RQ4)}
  \label{fig:KNLS_ratio}
\end{figure}

To simulate real-world log anomaly detection scenarios in which data accumulates in real-time, our model was tested using a limited number of known normal log sequences. Through this controlled sampling approach, we can assess whether a small number of known normal log sequences can provide robust performance. Figure~\ref{fig:KNLS_ratio} shows the performance when sampling a portion of the normal logs in each dataset, simulating a situation where only a small number of normal logs are known relative to the continuously accumulating test logs. From the results, RAPID shows stable and robust performance from using very little normal data to using most of the data. This means that our model can continue to detect in real-world log anomaly detection scenarios regardless of the accumulation of data, demonstrating that RAPID does not cause a training delay.

Such robust results can be explained by the following reasons. First, as outlined in Section~\ref{sec:log_token}, logs possess a limited vocabulary size, which means that even if a new test log is entered, the tokens that comprise it are most likely already included in the known normal log sequence. This characteristic gives R-LogAD the ability to interpret unseen test logs, and in particular because RAPID directly reflects all tokens, it provides in-depth understanding of unseen logs. More detailed analysis of the effect of utilizing all token information is performed in Appendix~\ref{sec:appendix_all_token}. Furthermore, even if we do not possess the exact same vocabulary of tokens for an unseen log, our methods can still interpret it based on words with similar meanings that are close in terms of PLM representation.

\section{Conclusion}
In this study, we propose RAPID, a model that improves the limitations of previous deep learning-based log anomaly detection methodologies through detailed analysis of log data. First, we reformulated the task to improve the delay of previous studies that require training for each log data. In the reformulated R-LogAD, anomaly detection is performed by comparing the test log and its nearest normal type without any training, and in the process, information loss is minimized by replacing the parameters of the log with regular expressions. Furthermore, noting that the pre-processed log sequences consist of only a common language, we extract the representation of each log sequence through a publicly available PLM. Second, RAPID uses a maxSim score that actively reflects the information of all tokens to enable more sophisticated detection and robustness against unseen test logs. Finally, we apply the core set technique, which requires minimal computation for realistic inference, to enable real-time detection.

To validate our proposed RAPID, we conducted experiments on three of the most commonly used benchmark datasets: BGL, Thunderbird, and HDFS. We addressed various research questions and found that our methodology demonstrated superior detection performance even without any training on log data. In particular, even in the unsupervised setting, where no information about abnormal logs is given, RAPID achieved competitive performance compared to models in the supervised setting, and the best performance on BGL. In addition to achieving efficient inference through the core set technique, it demonstrated robust performance in the log accumulation experiment, confirming its ability to deliver real-time results without training delays. Furthermore, even if new normal logs are continuously added, our model can be updated by simply including them in the known normal log sequence, making our approach highly scalable in the long run. Since RAPID is not limited to specific log data and can be directly applied to various computer systems, it is expected to be widely used in real-world industrial applications.

\section*{Acknowledgements}
We are grateful to Kyoungchan Park for his insightful feedback that helped improve this paper.

\bibliography{custom}
\bibliographystyle{acl_natbib}

\appendix
\section{Experiment setup}\label{sec:exp_setup}
\subsection{Dataset}
In this study, we choose BGL \cite{oliner2007supercomputers}, Thunderbird \cite{oliner2007supercomputers}, and HDFS \cite{xu2009detecting} log datasets as benchmarks to provide a fair comparison with previous work. All three datasets contain labels for normal and abnormal and are generated from different computer systems, allowing us to evaluate the generalization performance of RAPID. The BGL dataset contains logs from the Blue Gene/L supercomputer, each of which comes with a label separated by the alert category tag. The Thunderbird dataset was collected from the Thunderbird supercomputer operated by Sandia National Laboratories in Albuquerque, and it also contains alert and non-alert messages, similarly labeled with category tags. HDFS, on the other hand, is a log dataset derived from the Hadoop Distributed File System, where multiple logs output from a private cloud environment are grouped into blocks to form a single log instance. In BGL and Thunderbird, each single log is a data instance and is labeled, whereas HDFS has a label for each aggregated block of logs. Since our methodology is designed by analyzing single log sequences and assumes that queries and documents are single log sequences, BGL and Thunderbird are the target log datasets for RAPID, and the HDFS dataset is included as a supplemental dataset to check the robustness of the study. For BGL and HDFS, we used the full dataset, while for Thunderbird, we sample a portion of the dataset following the settings in \citet{9534113}~\footnote{https://github.com/HelenGuohx/logbert}, but maintained abnormal proportions. Since no training takes place in RAPID, we refer to known normal logs rather than training data. For BGL and Thunderbird datasets, we allocate 0.8 of the total normal logs as known normal logs and 0.2 as test normal logs, and all abnormal logs are included in the test logs. In HDFS, the same ratio is applied on a per-block basis to separate them. Details about the distribution of known normal logs and the log sequences (or blocks) used in the test dataset can be found in Table \ref{tab:Dataset}.

\begin{table} 
\centering
\resizebox{\columnwidth}{!}{
\setlength{\tabcolsep}{5pt}
\begin{tabular}{cccc} 
\Xhline{3\arrayrulewidth}
\textbf{Dataset} & \textbf{Type} & \textbf{Known Normal Log} & \textbf{Test}\\ \hline
\multirow{2}{*}
{\textbf{BGL}} & normal & 
\begin{tabular}[c]{@{}c@{}}3,519,602\end{tabular} & 
\begin{tabular}[c]{@{}c@{}}879,910\end{tabular}  \\ 
\cline{2-4}  & abnormal & - & \begin{tabular}[c]{@{}c@{}}348,460\end{tabular}\\ \hline
\multirow{2}{*}{\textbf{Thunderbird}}  & normal & 3,938,483 & 984,621   \\ 
\cline{2-4} & abnormal & - & 76,895 \\ \hline
\multirow{2}{*}{\textbf{HDFS}}  & normal & 446,578 & 111,645   \\ 
\cline{2-4} & abnormal & - & 16,838 \\ 
\Xhline{3\arrayrulewidth}
\end{tabular}}
\caption{Number of logs in each dataset used in RAPID}
\label{tab:Dataset}
\end{table}

\subsection{Baselines}
\label{sec:baselines}

In this study, we conduct comparative experiments with various existing models to evaluate and validate the performance of RAPID. The selected comparison models are categorized according to whether they use log-specific training, whether they use a parser, and how they utilize token information. Among the various benchmark models, the deep learning-based models are described below. 

\textbf{LogRobust} is a supervised learning model utilizing an attention-based bi-LSTM structure. It uses a log parser to pre-process the log data and generates TF-IDF and word semantic vectors to extract the features of the log data.

\textbf{HitAnomaly} is a supervised model that leverages a transformer structure. It uses a log parser to convert log data into standardized templates and encode log information into parameters.

\textbf{LogSy} is a supervised anomaly detection model that leverages a transformer structure. Log are pre-processed by a tokenizer and do not require the use of a log parser.

\textbf{Adanomaly} detects anomalies with feature extraction and ensemble methods using BiGAN model to address the class imbalance problem in log anomaly detection.

\textbf{DeepSyslog} suggests a way to represent the context of log events and event metadata. Based on the continuity of the log stream, it extracts the semantic and contextual information from logs using unsupervised learning sentence embedding.

\textbf{LogPal} combines template sequences and raw log sequences to generate log pattern events, automatically recognizing log patterns. This model performs binary classification.

\textbf{LayerLog} is a log anomaly detection framework based on the hierarchical semantics of log data. LayerLog effectively extracts semantic features from each layer and utilizes the word-sequence hierarchy.

\textbf{DeepLog} is a deep learning unsupervised log anomaly detection model based on LSTM structure. It utilizes a log parser to generate inputs for the LSTM and operates by predicting the next word.

\textbf{LogAnomaly} is a solution for detecting anomalies in log streams. By attention-based LSTM structure, it extracts semantic information from log templates using the template2vec technique.

\textbf{LogBERT} is a BERT-based anomaly detection model that utilizes MLM and DeepSVDD loss. After pre-processing with a log parser, it identifies anomaly patterns.

\textbf{LAnoBERT} is a BERT-based anomaly detection model. After training using MLM, it detects token-level anomalies by using MLM probability as an abnormal score in inference.

\subsection{Evaluation Metric} 

The threshold-dependent F1 score and threshold-independent AUROC are adopted as the evaluation metrics in this study. Once the threshold of the model is determined in the anomaly detection, the recall and precision are calculated through equation (\ref{eq: recall}) and equation (\ref{eq: precision}) depending on the actual anomaly and whether the anomaly is determined by the model, and the F1 score is calculated as the harmonic mean of these two indicators as shown in equation (\ref{eq: F1}).
\begin{align}
\text{Recall} &= \frac{TP}{TP+FN}, \label{eq: recall} \\
\text{Precision} &= \frac{TP}{TP+FP}, \label{eq: precision} \\
\text{F1 score} &= 2 \cdot \frac{\text{Precision} \cdot \text{Recall}}{\text{Precision} + \text{Recall}}. \label{eq: F1}
\end{align}
(TP: true positive, FP: false positive, FN: false negative.)

AUROC is a metric that calculates the false positive rate (FAR) and true positive rate (TPR) for all possible threshold candidates, then plots a receiver operating characteristic curve with FAR on the x-axis and TPR on the y-axis, and calculates the area under the curve. A better anomaly detection model will have a value of AUROC closer to 1, while a random model will have a value closer to 0.5.

In general, anomaly detection studies use AUROC as an evaluation metric to determine the intrinsic performance of a model that does not rely on a threshold, but existing studies in log anomaly detection utilize Best F1 score to measure the performance of classifying normal and abnormal logs. Moreover, since studies in an unsupervised setting, including ours, only consider normal data, the Best F1 Score threshold cannot be predetermined in advance. Therefore, we compute the Best F1 Score in the same way as previous studies, using the threshold that gives the best theoretical performance on the test dataset. In addition, most of the previous studies except LAnoBERT do not report the performance of AUROC score, so we cannot compare the AUROC of baseline models together. The AUROC performance of RAPID is recorded in \ref{tab:plm}.

\subsection{Evaluation Details} 
All evaluation experiments are performed on a single Linux system, and the specific environment is as follows. Four TITAN-RTX GPUs are used to extract the PLM representation of log sequences during database construction, one TITAN-RTX GPU is used to compute \textsf{maxSim} between query-documents, and all other processes are performed on the CPU. The PLM utilized the Huggingface Transformers library \citep{wolf-etal-2020-transformers} for implementation. The methodology does not impose strict restrictions on the choice of PLM, and except for the experiments in Section~\ref{sec:plm} which examine the dependency on a specific PLM, all experiments employ BERT to obtain a representation of the log sequence. The proposed RAPID is a model that does not perform any training and therefore does not have any training parameters. When processing the data through the PLM, we set the max token length to 128 for the BGL and Thunderbird datasets and 512 for the HDFS dataset. This difference is due to the fact that the data in each type is labeled at the log sequence level or at the log block level, respectively, so we set a relatively long max length for HDFS, which consists of one input per block. In addition, in the case of HDFS, to ensure that the input per block does not exceed the maximum token length of BERT, the input is constructed by concatenating only unique log sequences within the block. Finally, the only hyperparameter, the size of the core set, determines how much of the completed $D$ for each dataset is taken as the core set. In Section~\ref{sec:real_time}, we verify that the final anomaly detection performance is robust to the size of the core set, and finally set it to utilize only 0.01 unique known normal log sequence out of the total log sequences in $D$.

\section{Additional Research Questions}\label{sec:additional_RQ}

\subsection{Is a general-purpose PLM enough to detect log anomalies?}\label{sec:plm}
To demonstrate that the performance of RAPID is not dependent on a specific PLM model, we compared its performance using three well-known general-purpose PLMs: BERT, RoBERTa \cite{liu2019roberta}, and ELECTRA \cite{clark2020electra}.
\begin{table}
\centering
\resizebox{\columnwidth}{!}{%
\setlength{\tabcolsep}{5pt}
\renewcommand{\arraystretch}{1.4}
\begin{tabular}{l|c|c|c|c|c|c} 
\Xhline{3\arrayrulewidth}
\multirow{2}{*}{\textbf{PLM}} & \multicolumn{2}{c|}{\textbf{BGL}} & \multicolumn{2}{c|}{\textbf{Thunderbird}} & \multicolumn{2}{c}{\textbf{HDFS}}  \\ 
\cline{2-7}
 & \textbf{F1} & \textbf{AUROC}      & \textbf{F1} & \textbf{AUROC}              & \textbf{F1} & \textbf{AUROC}       \\ 
\hline
BERT               & 0.9999      & 0.9999              & 0.9979      & 0.9994                      & 0.9240      & 0.9295               \\
RoBERTa                      & 0.9999      & 0.9999              & 0.9968      & 0.9992                      & 0.9249      & 0.9303               \\
ELECTRA        & 0.9999      & 0.9999              & 0.9982      & 0.9992                      & 0.9240      & 0.9295               \\
\Xhline{3\arrayrulewidth}
\end{tabular}}
\caption{Performance based on the general purpose PLM used. Robust performance was observed across all three well-known PLMs we tested.}
\label{tab:plm}
\end{table}

Table~\ref{tab:plm} shows the performance of utilizing well-known PLMs BERT, RoBERTa, and ELECTRA to obtain the log sequence representation. All three PLMs show robust performance, indicating that R-LogAD can be accomplished by utilizing general-purpose PLMs without any additional training. Incidentally, to ensure a fair comparison with LogBERT and LAnoBERT, all experiments recorded performance based on BERT.

In our method, we use a \textsf{maxSim} distance that actively reflects all token information in the query and document log sequences when performing R-LogAD. Based on the analysis in Section~\ref{sec:log_token}, this approach is intended to provide a deeper interpretation of unseen test logs, and we expect that the effectiveness of this approach will be more pronounced as the amount of unseen test logs increases. Therefore, we conduct an experiment with increasing the proportion of unseen test logs by decreasing the proportion of known normal log sequences and comparing the performance of using only sequence-level information of CLS token and using information of all tokens. The results are shown in Table~\ref{tab:effect_all_token}.

\subsection{Is it effective to consider all tokens?}
\label{sec:appendix_all_token}
\begin{table}
\centering
\resizebox{\columnwidth}{!}{%
\begin{tabular}{c|c|r|cc}
\hline
\Xhline{3\arrayrulewidth}
\multirow{2}{*}{\textbf{Dataset}} & \multirow{2}{*}{\textbf{vocabulary size}} & \multicolumn{1}{c|}{\textbf{known normal}} & \multicolumn{2}{c}{\textbf{F1}} \\ \cline{4-5}
                                  &                                      & \multicolumn{1}{c|}{\textbf{log sequence ratio}}  & \multicolumn{1}{c|}{\textbf{CLS only}} & \textbf{all token} \\ \hline
\multirow{4}{*}{\textbf{BGL}}        & \multirow{4}{*}{888}  & 1                     & \multicolumn{1}{c|}{0.9999}             & 0.9999              \\
                                     &                       & 0.1                   & \multicolumn{1}{c|}{0.9999}             & 0.9999              \\
                                     &                       & 0.01                  & \multicolumn{1}{c|}{0.9994}             & 0.9988              \\
                                     &                       & 0.001                 & \multicolumn{1}{c|}{0.9946}             & 0.9924              \\ \hline
\multirow{4}{*}{\textbf{Thunderbird}} & \multirow{4}{*}{3,137} & 1                     & \multicolumn{1}{c|}{0.9972}             & 0.9979              \\
                                     &                       & 0.1                   & \multicolumn{1}{c|}{0.9886}             & 0.9941              \\
                                     &                       & 0.01                  & \multicolumn{1}{c|}{0.9785}             & 0.9959              \\
                                     &                       & 0.001                 & \multicolumn{1}{c|}{\textbf{0.7884}}    & 0.9690              \\ \hline
\Xhline{3\arrayrulewidth}
\end{tabular}}
\caption{Comparison between using only CLS and all tokens. From Thunderbird results, we can see that reflecting information from all tokens contributes to robust performance as fewer known normal log sequences result in more unseen test logs.}
\label{tab:effect_all_token}
\end{table}

For the Thunderbird dataset, as the percentage of unseen test logs increased, the performance of utilizing all token information shows more robustness. On the other hand, for the BGL dataset, utilizing all token information does not show a noticeable performance difference, which is likely due to the very small vocabulary size of the dataset itself (888), where the CLS token is sufficient to summarize the overall meaning of the logs. Nevertheless, even for log datasets such as BGL, as the number of log types increases due to continuous data accumulation and system updates, the vocabulary size is expected to increase\cite{10.1109/ASE51524.2021.9678773, 9865986}, and the utilization of all token information is expected to become increasingly important.

\end{document}